\documentclass[letterpaper, 10 pt, conference]{ieeeconf}  

\IEEEoverridecommandlockouts                              

\usepackage{graphicx}
\usepackage[normalem]{ulem}
\usepackage{amsmath,amssymb,amsfonts}
\usepackage{float}
\usepackage{booktabs}
\usepackage{multirow}
\usepackage[tight,footnotesize]{subfigure}

\newcommand{\bftheta}{\boldsymbol\phi}
\DeclareMathOperator*{\argminA}{arg\,min}

\title{\LARGE \bf
Leveraging Predictive Models for Adaptive Sampling of\\Spatiotemporal Fluid Processes
}

\author{Sandeep Manjanna$^{1}$, Tom Z. Jiahao$^{2}$, and M. Ani Hsieh$^{2}$
\thanks{*This work was not supported by any organization}
\thanks{$^{1}$ Plaksha University, Mohali, India. {\tt\footnotesize \{msandeep.sjce@gmail.com\}}}
\thanks{$^{2}$ GRASP Laboratory, University of Pennsylvania, Philadelphia, USA.
        {\tt\footnotesize \{zjh, m.hsieh\}@seas.upenn.edu}}
}

\begin{document}

\maketitle

\begin{abstract}
Persistent monitoring of a spatiotemporal fluid process requires data sampling and predictive modeling of the process being monitored. In this paper we present \textbf{PASST} algorithm: \textbf{P}redictive-model based \textbf{A}daptive \textbf{S}ampling of a \textbf{S}patio-\textbf{T}emporal process. PASST is an adaptive robotic sampling algorithm that leverages predictive models to efficiently and persistently monitor a fluid process in a given region of interest. Our algorithm makes use of the predictions from a learned prediction model to plan a path for an autonomous vehicle to adaptively and efficiently survey the region of interest. In turn, the sampled data is used to obtain better predictions by giving an updated initial state to the predictive model. For predictive model, we use Knowledged-based Neural Ordinary Differential Equations to train models of fluid processes. These models are orders of magnitude smaller in size and run much faster than fluid data obtained from direct numerical simulations of the partial differential equations that describe the fluid processes or other comparable computational fluids models. For path planning, we use reinforcement learning based planning algorithms that use the field predictions as reward functions. We evaluate our adaptive sampling path planning algorithm on both numerically simulated fluid data and real-world nowcast ocean flow data to show that we can sample the spatiotemporal field in the given region of interest for long time horizons. We also evaluate \textbf{PASST} algorithm's generalization ability to sample from fluid processes that are not in the training repertoire of the learned models.
\end{abstract}

\section{Introduction}

In this paper we present \textbf{PASST} algorithm: \textbf{P}redictive-model based \textbf{A}daptive \textbf{S}ampling of a \textbf{S}patio-\textbf{T}emporal process. PASST is an adaptive robotic sampling algorithm that leverages predictive models to efficiently and persistently monitor a fluid process in a given region of interest. Our algorithm makes use of the predictions from a learned prediction model to plan a path for an autonomous vehicle to adaptively and efficiently survey the region of interest. In turn, the sampled data is used to obtain better predictions by giving an updated initial state to the predictive model.

\begin{figure}[ht]
    \begin{center}
    \includegraphics[width=0.95\columnwidth]{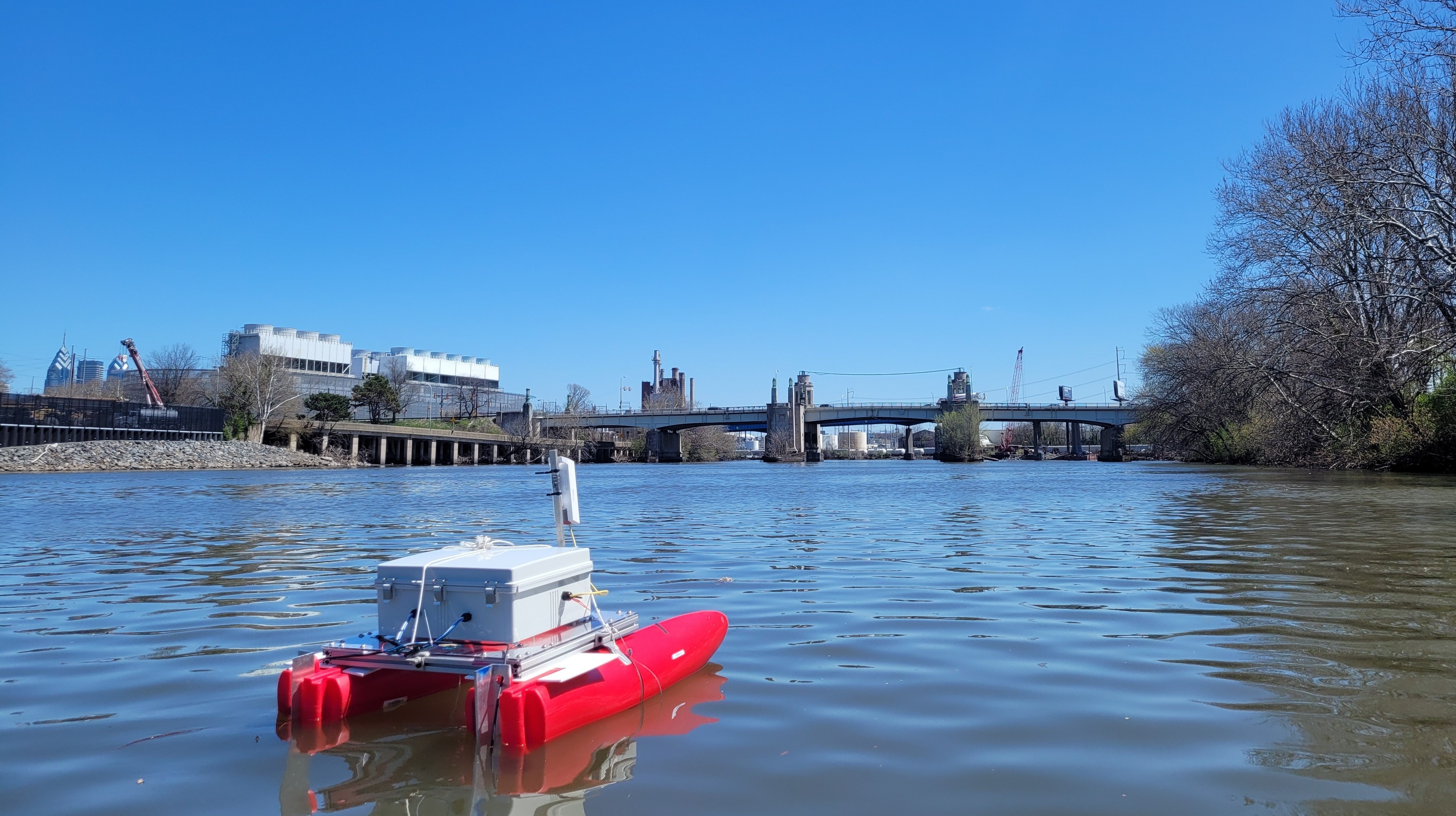}
    \end{center}
    \caption{An example application of using autonomous surface vehicles to sample water quality measurements from a river.}
    \vspace{-1.5em}
    \label{fig:boat}
\end{figure}

Understanding the interplay between data acquisition and modeling of dynamics in environmental phenomena is crucial for learning about many spatiotemporal physical processes. One of the important pieces slowing down the efficient tackling of grand issues, such as climate change or non-sustainable agriculture practices, is the lack of understanding of physical environmental phenomena at larger scale. Currently, far less is known about many large-scale environmental processes. For example, even though we are very reliant on the ocean for driving weather, regulating temperature, and ultimately supporting all living organisms, more than $80\%$ of this vast, underwater realm remains unmapped, unobserved, and unexplored~\cite{noaa}. We envision various autonomous systems monitoring, sampling, and collecting information about our environment and helping in bridging the gap between current environmental models and the underlying input data. To achieve such persistent monitoring systems, we propose an algorithm to adaptively sample a spatiotemporal process using an autonomous vehicle. Fig.~\ref{fig:boat} presents an example application of using autonomous surface vehicles to sample water quality measurements from Schuylkill river in Philadelphia.

From a robotics perspective, it is important to answer these two questions --- 1. How are the actions of the robotic system enabling a better understanding of the environment? 2. How can one use the inferred environmental models to inform the controls and design of a robotic system? As noted by~\cite{doi:10.1146/annurev-fluid-010719-060214}, the lack of consideration for fluid processes in the design of aquatic robots may impact their performance. In this paper we are interested in answering the these questions and developing techniques to enable robots to selectively survey an environment to map a spatiotemporal process over time. The need for adaptive sampling comes from the fact that the processes being monitored are temporal and such adaptive surveys also help in extending our approach to large-scale environments. The key contributions of this paper include:

\begin{itemize}
    \item Knowledge-based Neural Ordinary Differential Equation (KNODE) predictive models that are orders of magnitude smaller in size than fluid data, and run much faster than solving PDEs, thus are suitable for deployment on robots.
    \item Reinforcement learning based adaptive robotic sampling algorithm that leverages predictive models to efficiently and persistently monitor a spatiotemporal fluid process in a given region of interest. 
\end{itemize}

As we are sampling from a spatiotemporal process, we consider both Root Mean Squared Error (RMSE or MSE) and Proper Orthogonal Decomposition (POD) metrics to evaluate our approach. RMSE or MSE between the ground truth and temporal estimates
of the fluid process provides an insight into the spatial similarity index captured in the samples collected. POD compares the energy distribution between the estimated flow and the ground truth process, thus capturing the spatiotemporal similarity index.

\section{Related Work}
In recent decades, there has been increasing focus on reduced-order modeling of spatiotemporal processes to describe their features in more compact and efficient manners. Prominent approaches include POD \cite{holmes_lumley_berkooz_1996}, which is based on singular value decomposition, and identifies orthogonal modes from a fluid process to embed high-dimensional fluid data in a low-dimensional space. Other approaches include dynamic mode decomposition (DMD)~\cite{doi:10.1146/annurev-fluid-030121-015835} and Koopman operators~\cite{koopman}, which are closely related: DMD extracts low-dimensional linear models of coherent structures from fluid processes; Koopman operator models the evolution of measurement functions by lifting the fluid process to an infinite-dimensional space. In addition, sparse regression techniques such as SINDy~\cite{SINDy} and entropic regression~\cite{entropicregression} were proposed to build parsimonious fluid models, but suffer from the need for a predetermined library of candidate functions. In the field of deep learning, recurrent neural networks have been used to model high-dimensional flows \cite{qraitem2020bridging} and spatiotemporally chaotic processes \cite{OttRescomp}. Physics-informed neural networks have enabled the incorporation of physical constraints into solving PDEs \cite{physicslearning}. Most notably, Chen et al. proposed neural ordinary differential equations (NODE), which is a new family of continuous-depth residual networks \cite{conf/nips/ChenRBD18}. NODE and its variants are a natural fit for modeling continuous-time processes and has demonstrated its ability to model a variety of high-dimensional dynamical systems \cite{wu2022nodeo, NPDE, Jiahao2021Knowledgebased}. In particular, convolutional layers, graph filters, and physical constraints can be readily incorporated into NODE models to improve training \cite{wu2022nodeo, zang2020neural, duong2021hamiltonianbased}. In this work, we will use the variant KNODE to train models of fluid processes by incorporating simple inductive hypothesis.

Spatiotemporal process sampling has been the focus in many real-world tasks such as feature mapping and process monitoring. Applications include mapping of relatively stationary objects such as coral reefs~\cite{diverfollowingDas} and geologic features~\cite{chen2021autonomous}, and monitoring of dynamic processes such as ocean flows~\cite{tahiya1}, oil spills~\cite{PASHNA2020107238}, and dissolved green house gases~\cite{nicholson2018rapid}. Sampling can benefit from predictive models to improve robots' sampling actions. Reduced-order modeling methods such as POD have been use to inform robot teams about sampling locations for minimizing flow field reconstruction errors~\cite{tahiya1}, and Koopman operators have been used to extract key features from flows to improve the situational awareness of robots~\cite{tahiya2}. There is a large literature on robotic data sampling and coverage, much of which is based on mechanisms that use waypoints of geometric priors with limited dependence on the distribution of incoming observations~\cite{pizarro, Kemna-2018-986}. Compared to complete coverage algorithms~\cite{choset1996sensor}, adaptive sampling approaches trade off completeness for efficiency. In many cases, even if the process being studied is rapidly varying, sub-sampling can be effective when the sample points are correctly selected~\cite{venkataramani2000perfect}. If the environment contains peaks with high local-variance, adaptive sampling can exploit the clustering phenomena to map the environmental field more accurately than non-adaptive sampling~\cite{manjanna2017data}. Our sampling approach places emphasis on the density of valuable measurements that can be collected in a limited amount of time and also gets continuous inputs from a predictive model to achieve persistent monitoring of a spatiotemporal fluid process.

\begin{figure}[ht]
    \begin{center}
    \includegraphics[width=\columnwidth]{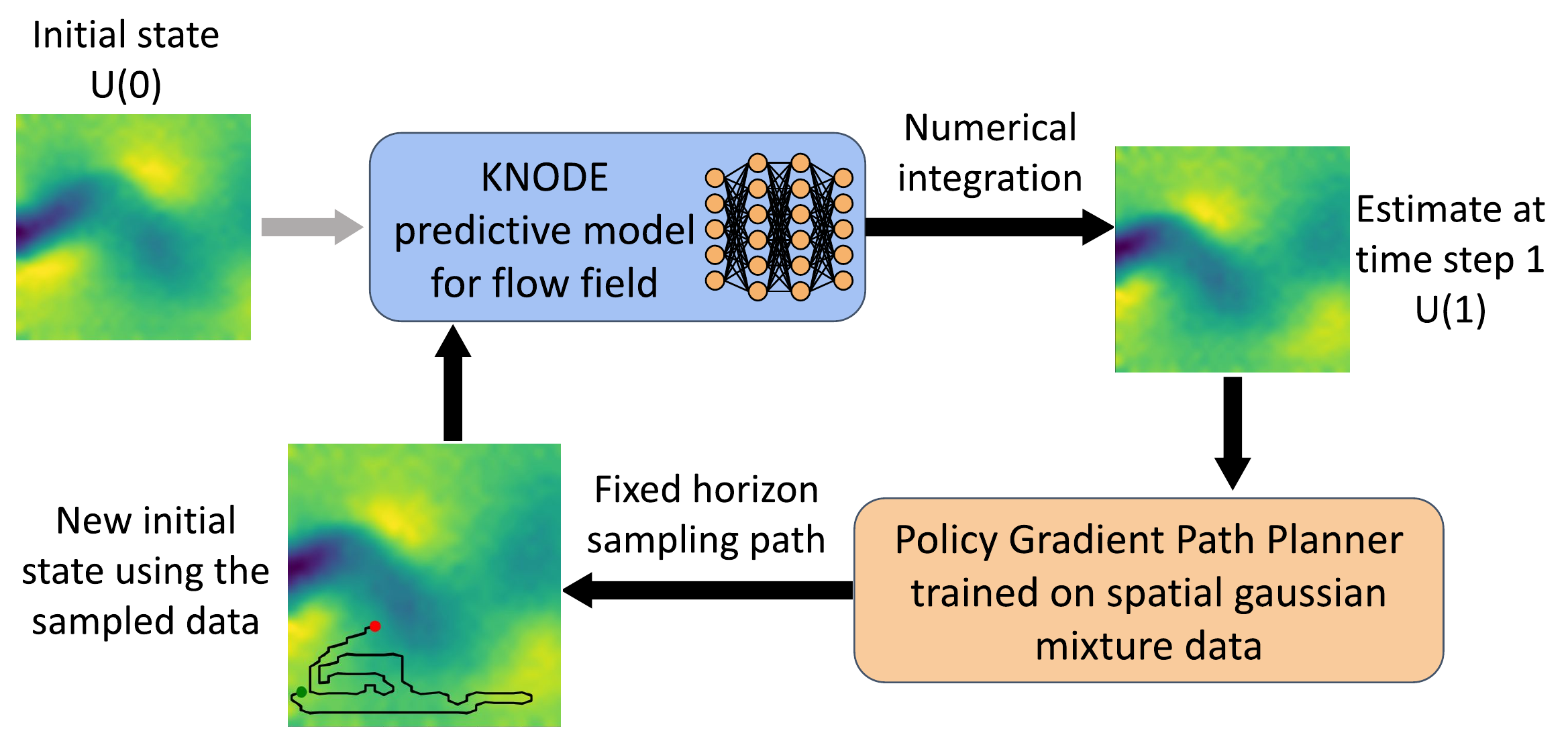}
    \end{center}
    \caption{\textbf{PASST Overview:} This figure presents an overview of our approach where the KNODE predictive model together with Policy gradient based path planning generates policies for a sampling robot to persistently survey a spatiotemporal fluid process.}
    \label{fig:overview}
\end{figure}

\begin{figure*}[ht]
    \begin{center}
    \vspace{1 em}
    \includegraphics[width=\textwidth]{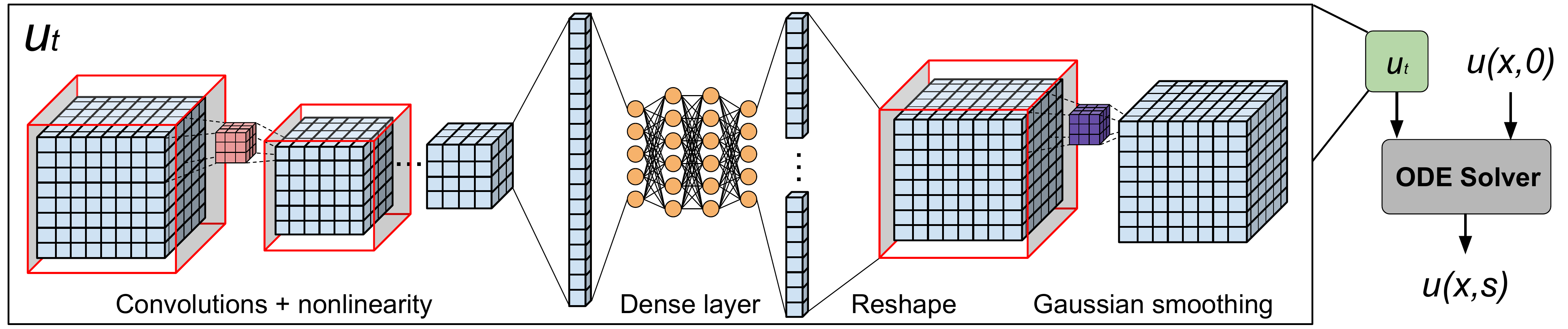}
    \end{center}
    \caption{\textbf{KNODE architecture.} The input snapshot of the flow data first gets down-sampled using convolutional layers, and then get passed through dense layers. Finally it gets reshaped back to the original dimension, and smoothed with Gaussian kernels at the output.}
    \label{fig: node arch}
\end{figure*}

\section{Problem Formulation}

The sampling region is a continuous two-dimensional area of interest $\mathcal{E} \subset \mathbb{R}^2$ with user-defined boundaries. We consider a discretized region with uniform grid cells such that the robot's position $x$ can be represented by a pair of integers $x \in \mathbb{Z}^2$. Each grid cell $(i,j)$ has an associated value or information or reward $q(i,j)$ that the robot will gain when a measurement is taken in that grid cell~\cite{manjanna2018reinforcement}. But, the process being sampled is spatiotemporal, hence the reward $q$ is also a function of time. Thus, each grid cell $(i,j)$ has an associated value $q(i,j,t)$, where $t$ is for time. Given a bounded region of interest, the objective is to efficiently and persistently monitor a spatiotemporal process by collecting measurements with an autonomous vehicle.

\section{Approach}

Let $\vec{x}$ be the list of all the locations $x \in \mathbb{Z}^2$ visited by the sampling robot in a fixed sampling horizon. The above mentioned objective of persistently monitoring a spatiotemporal process can be mapped to maximizing the total accumulated reward $J$ by the robot sampler over a fixed time horizon $T$. 
\begin{equation}
    J = \sum_{t=0}^{t=T} q(i,j,t)\hspace{1cm}\forall (i,j) \in \vec{x}
\end{equation}

This is a problem of sampling a spatiotemporal process and our solution splits this problem into two pieces ---
\begin{enumerate}
    \item Fluid field modeling to generate predictions about how the process evolves over temporal space.
    \item Adaptive sampling of a spatial field to achieve efficient and selective survey of the process.
\end{enumerate}

These two components, a predictive model for spatiotemporal fluid process and a rewardmap based policy gradient robot path planner, combined together form our proposed algorithm \textbf{PASST}: \textbf{P}redictive-model based \textbf{A}daptive \textbf{S}ampling of a \textbf{S}patio-\textbf{T}emporal process. Fig.~\ref{fig:overview} presents an overview of the \textbf{PASST} algorithm. The trained KNODE predictive model takes an initial state as input and predicts the evolution of the fluid process over time. The predictions from the KNODE model are more accurate for the immediate time steps and start to degrade over long periods of time in future. This is not ideal for persistent monitoring. Hence, as illustrated in Fig.~\ref{fig:overview}, we use the one-step predictions from the KNODE model to generate adaptive sampling path for the robot. Once fixed horizon measurements of the field are collected, a spatial snapshot of the field is created with the acquired samples and used as an initial state on the predictive model to predict the further evolution of the process in the future time steps. Thus the planner algorithm leverages the predictive model and in turn the predictive model is getting an updated initial state. This way, a spatiotemporal fluid process can be persistently monitored. We make an assumption that the process being monitored is changing slowly compared to the speed of a robotic sampler. In the next subsections we will discuss the details of both the KNODE predictive model and policy gradient based spatial sampler.

\subsection{Predictive Models for Spatiotemporal Processes}

Consider a spatiotemporal process $u(x, t) \in \mathbb{R}^n$, where the vector $u$ is the quantity of interest with respect to the process, the vector $x\in \mathbb{Z} \subset \mathbb{R}^m$ is a location in the environment $\mathbb{Z}$, and $t$ is time, we can describe the evolution of $u$ using a partial differential equation (PDE) given by
\begin{equation}
    u_t = \mathcal{N}(u, u_x, u_{xx},\cdots, x, t),
    \label{eqn: fluid process}
\end{equation}
where $u_t$ is the partial derivative of $u$ with respect to time $t$, and $u_x, u_{xx}, \cdots$ are the partial derivatives of $u$ with respect to space. Given past observations of the process $\mathcal{O} = [u(x,t_0), u(x,t_1), \cdots, u(x, t_p)]$ for all $x \in \mathbb{Z}$ over the time interval $\mathbb{T}_{\mathcal{O}} = [t_0, t_1, \cdots, t_p]$, our goal is build a model that can predict future values of $u$ for $\mathbb{T} = [t_{p+1}, t_{p+2},\cdots]$. In this work, we build such a model by approximating $\mathcal{N}$ using a neural network, which describes the evolution of the process as
\begin{equation}
    u_t = \hat{\mathcal{N}}(u; \bftheta),
    \label{eqn: NN general}
\end{equation}
where $\hat{\mathcal{N}}$ is a neural network parameterized by $\bftheta$. Since the process uses Eulerian description and the values of $u$ are observed at fixed locations, it can be said that the process is spatially discretized into a system of ODEs.

The neural network architecture of $\hat{\mathcal{N}}$ is shown in Fig.~\ref{fig: node arch}. While it is flexible to incorporate various physical priors into \eqref{eqn: NN general}, in this work we only apply Gaussian smoothing in the model, which results in the KNODE model given by

\begin{equation}
    u_t = \mathcal{K}\hat{\mathcal{N}}(u;\bftheta),
\end{equation}
where $\mathcal{K}$ denotes the spatial Gaussian smoothing. For training, we define a loss function given by
\begin{equation}
    \label{eqn: objective}
    L(\bftheta) = \frac{1}{|\mathbb{Z}|(|\mathbb{T_{\mathcal{O}}}|-1)}\sum_{x \in \mathbb{Z}}\sum_{\substack{t \in \mathbb{T}\\
    t \neq t_0}}\|\hat{u}(x, t) - u(x, t)\|_2^2,
\end{equation}
where $|\mathbb{Z}|$ and $|\mathbb{T}|$ are the size of the lattice grid and total number of observed time steps. The vector $\hat{u}(x, t)$ is the one-step-ahead model prediction using $u(x,t-1)$ as the initial condition. It is given by
\begin{equation}
    \hat{u}(x, t) = u(x, t-1) + \int^t_{t-1}\mathcal{K}\hat{\mathcal{N}}(u;\bftheta) dt.
    \label{eqn: one-step-ahead prediction}
\end{equation}
In practice, the integration in equation \eqref{eqn: one-step-ahead prediction} is performed using numerical solvers. Note that for the state at $t=0$, we use the symbol $U(0)$ to denote $u(x, 0), x\in\mathbb{Z}$.

Overall, given the observation data, we solve the optimization problem given by $\bftheta = \argminA_{\bftheta \in \Phi} L(\bftheta)$, where $\Phi$ is the space of all parameters. To solve this optimization task, we use the adjoint method similar to \cite{conf/nips/ChenRBD18} and \cite{Jiahao2021Knowledgebased}. Alternatives such as backpropagation could be used as well.

\subsection{Policy Gradient based Path Planning for Robots}

\begin{figure}[ht]
    \begin{center}
    \includegraphics[width=\columnwidth]{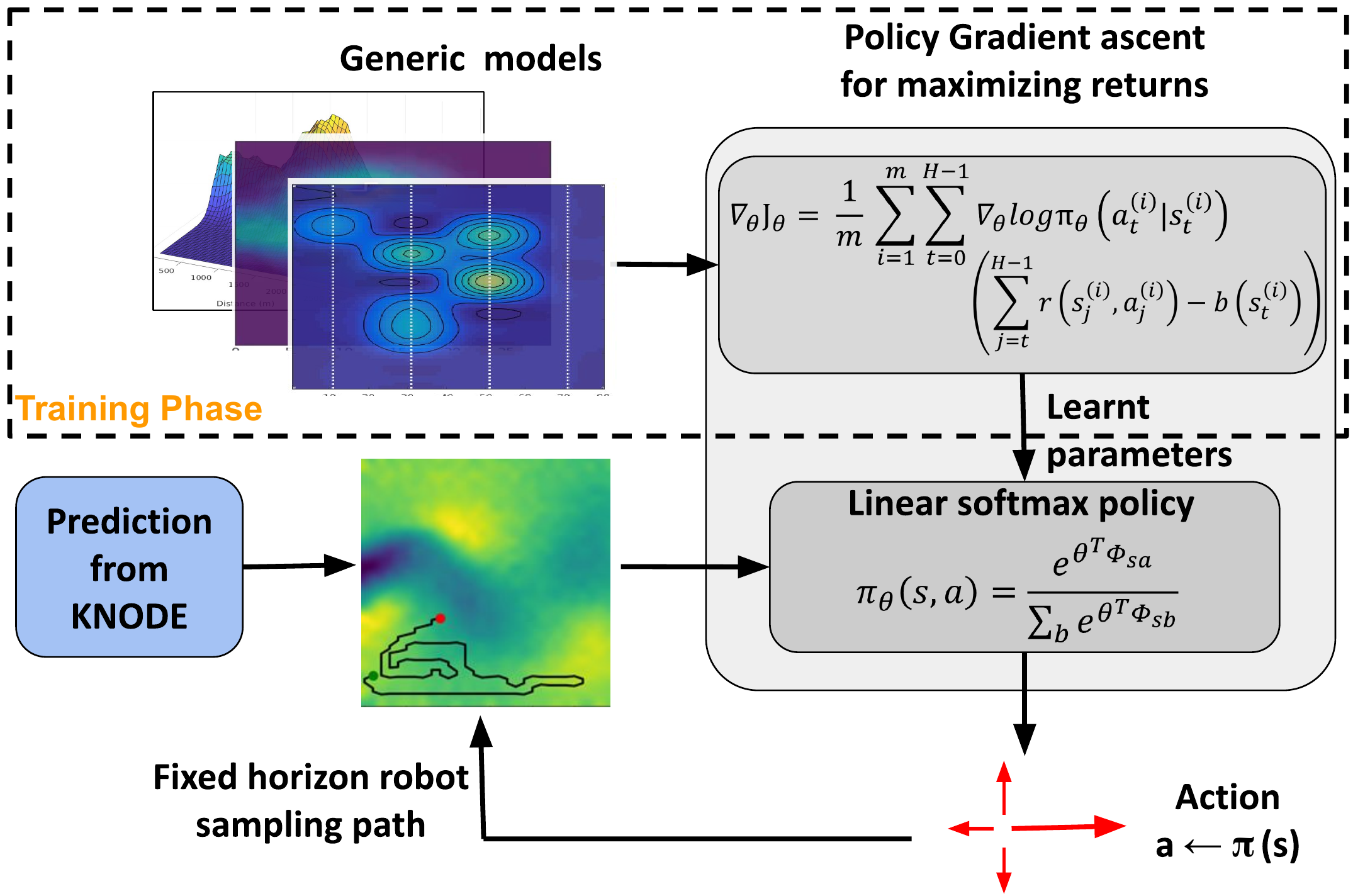}
    \end{center}
    \caption{\textbf{Policy Gradient based path planning:} An illustration of the training and test phase of the robotic path planning algorithm used in our adaptive sampling approach.}
    \label{fig:pg_overview}
\end{figure}

In our previous work~\cite{manjanna2022scalable}, we have formulated this problem as a Markov Decision Process and applied policy gradient search that directly optimizes the policy parameters $\boldsymbol{\theta}$ based on (simulated) experiences. This has been shown very efficient for sampling from a static spatial distribution. Fig.~\ref{fig:pg_overview} illustrates the use of a policy gradient approach to generate high-rewarding non-myopic sampling paths for the robot. In this paper, we team up this spatial sampling algorithm with a fluid-physics predictive model to generate sampling paths for a robot to achieve persistent monitoring of a spatiotemporal fluid process.

The policy gradient path planning approach illustrated in Fig.~\ref{fig:pg_overview} has a training phase where the policy parameters $\boldsymbol{\theta}$ are learnt using generic gaussian mixture models. Then, in the testing phase, a one-step prediction from the KNODE predictive model is taken as rewardmap to plan a sampling path for the robot in real-time. The measurements are collected along this path and are used to further generate a field map of the underlying spatiotemporal process. This field map acts as the next initial state $U(0)$ for the KNODE predictive model. Note that although KNODE has been used for online \textit{model} retraining and update in previous applications~\cite{jiahao2022online}, this work, we update the \textit{input} to the KNODE model instead of the model itself.

\section{Experiments and Results}
\begin{figure}
    \centering
    \vspace{1 em}
    \includegraphics[width=0.45\textwidth]{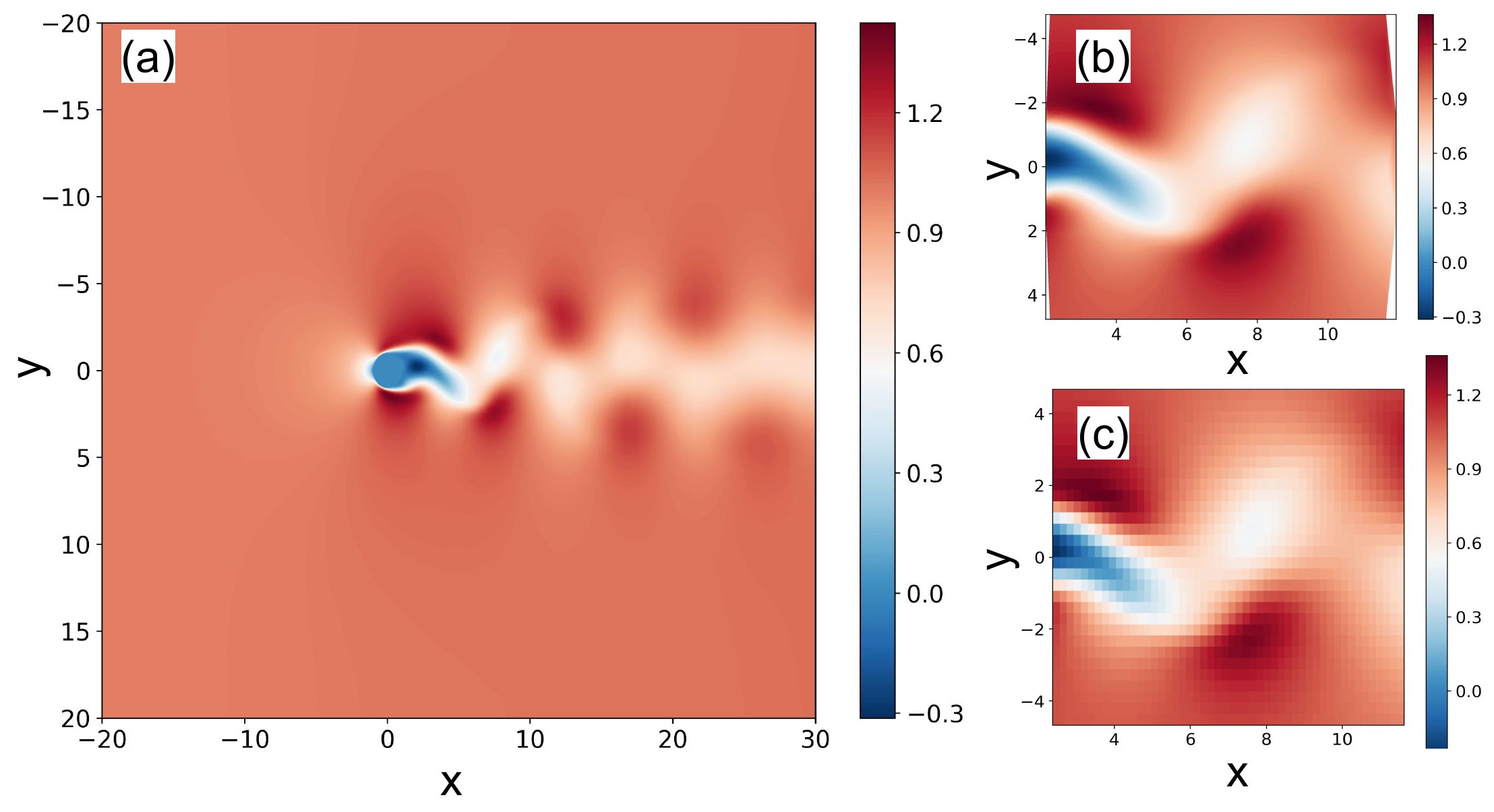}
    \caption{\textbf{Flow behind a cylinder.} The subfigures show (a) the simulated data in the environment, (b) cropped out subregion of the flow, and (c) processed data on a lattice grid. All snapshots are taken at the 200th step, where the vortex shedding has reached a steady state.}
	\vspace{-1em}
    \label{fig: vortex training data}
\end{figure}

We consider three different fluid processes to evaluate our approach. Two synthetic fluid flows generated through numerical simulations and one nowcast ocean data generated by National
Oceanic and Atmospheric Administration (NOAA). The first process is a 2D flow around a cylinder. The flow is simulated by solving the Navier-Stoke equations in a 50m by 40m rectangular work space. An inflow stream with uniform velocity profile is imposed on the left boundary, with a zero pressure outflow on the right boundary. The cylinder has a 1m radius and is centered at (0, 0). The Reynolds number of the flow is 200. For training data, we use a 10m by 10m subregion in the work space, and then process the data to be on a lattice grid using linear interpolation, which results in a 30 by 30 grid. The simulation work space, subregion, and processed training data are shown in Fig.~\ref{fig: vortex training data}. Vortex shedding is fully developed around the $200^{th}$ snapshots, and therefore the training data is the flow field from $200^{th}$ to $300^{th}$ snapshots, with a total of 100 steps. The $300^{th}$ to $400^{th}$ snapshots are used as the testing data.

The second process that we consider is a variant of the above mentioned vortex shedding 2D flow, but with the cylinder oscillating up and down at a frequency of one oscillation every 96 time steps. This process is used to validate the ability of \textbf{PASST} algorithm to generalize to sample from fluid processes that are not in the training repertoire. We do this by training only on the static cylinder vortex shedding data and testing the sampling performance on the vortex shedding process generated by oscillating cylinder.


We train a model using only the flow data with stationary cylinder. The model architecture is shown in Fig.~\ref{fig: node arch}. Specifically, we use five convolution layers, two linear layers, and a final Gaussian smoothing layer with kernel size 5, and variance 0.1. Besides, we added zero-mean Gaussian white noise with variance 0.001, which helps with stabilizing the model~\cite{OttRescomp}. We use the Adam optimizer with a learning rate of 0.001, and trained a total of 1981 epochs.

\subsection{KNODE Model Evaluation}\label{subsec:model_eval}
\begin{figure}[h]
    \centering
    \includegraphics[width=0.40\textwidth]{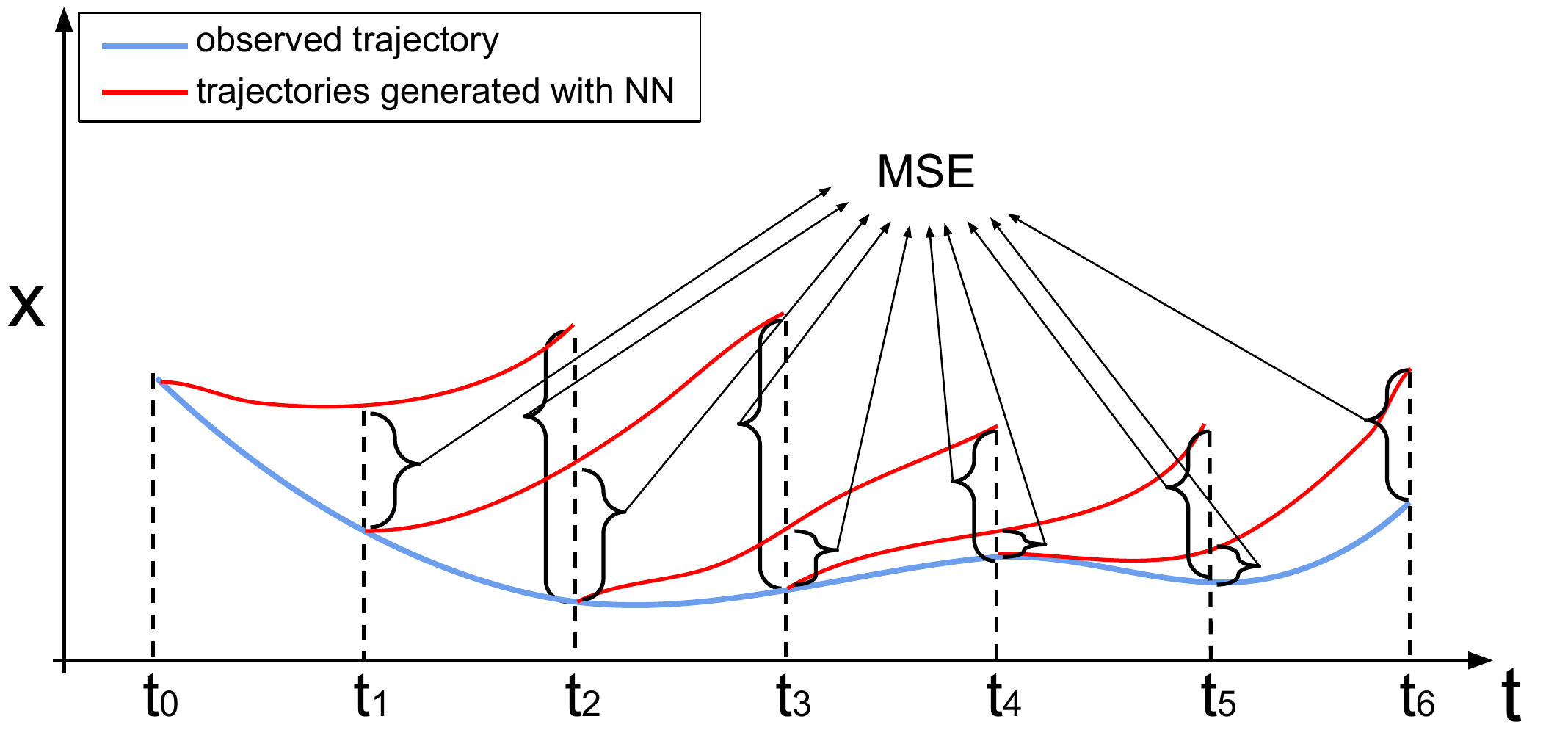}
    \caption{\textbf{Evaluation metrics.} An example of computing the total MSE with a lookahead of 2.}
	\vspace{-1em}
    \label{fig:eval_metrics}
\end{figure}
To evaluate the performance of the trained fluid model, we use both quantitative and qualitative metrics. Note that in this work, the goal is not to have a fluid model that is the most accurate, but to show that an imperfect fluid models, combined with adaptive sampling, can improve the performance of each other. 

For the quantitative metrics, we use the trained fluid model to perform prediction from a set of initial conditions for different durations, and then compute both the mean squared error (MSE) and average cosine distance between the predictions and ground truth. We use MSE to evaluate the learned flow field magnitude, while cosine distance is used to evaluate the direction of the velocity vectors, similar to the usage in~\cite{qraitem2020bridging}. Specifically, on the training, testing, and oscillating cylinder datasets, we use every snapshots in the data as the initial conditions to generate predictions with the learned model. The number of steps to predict from each initial conditions is called the \textit{lookahead}. This metric is illustrated in Fig.~\ref{fig:eval_metrics}.

For the qualitative metric, the goal is to evaluate whether the flow generated by the learned model is ``like'' the ground truth. To achieve this, We adopt POD to compare the energy distribution of the learned flow~\cite{holmes_lumley_berkooz_1996}. 
\begin{figure}[h]
    \centering
    \includegraphics[width=0.45\textwidth]{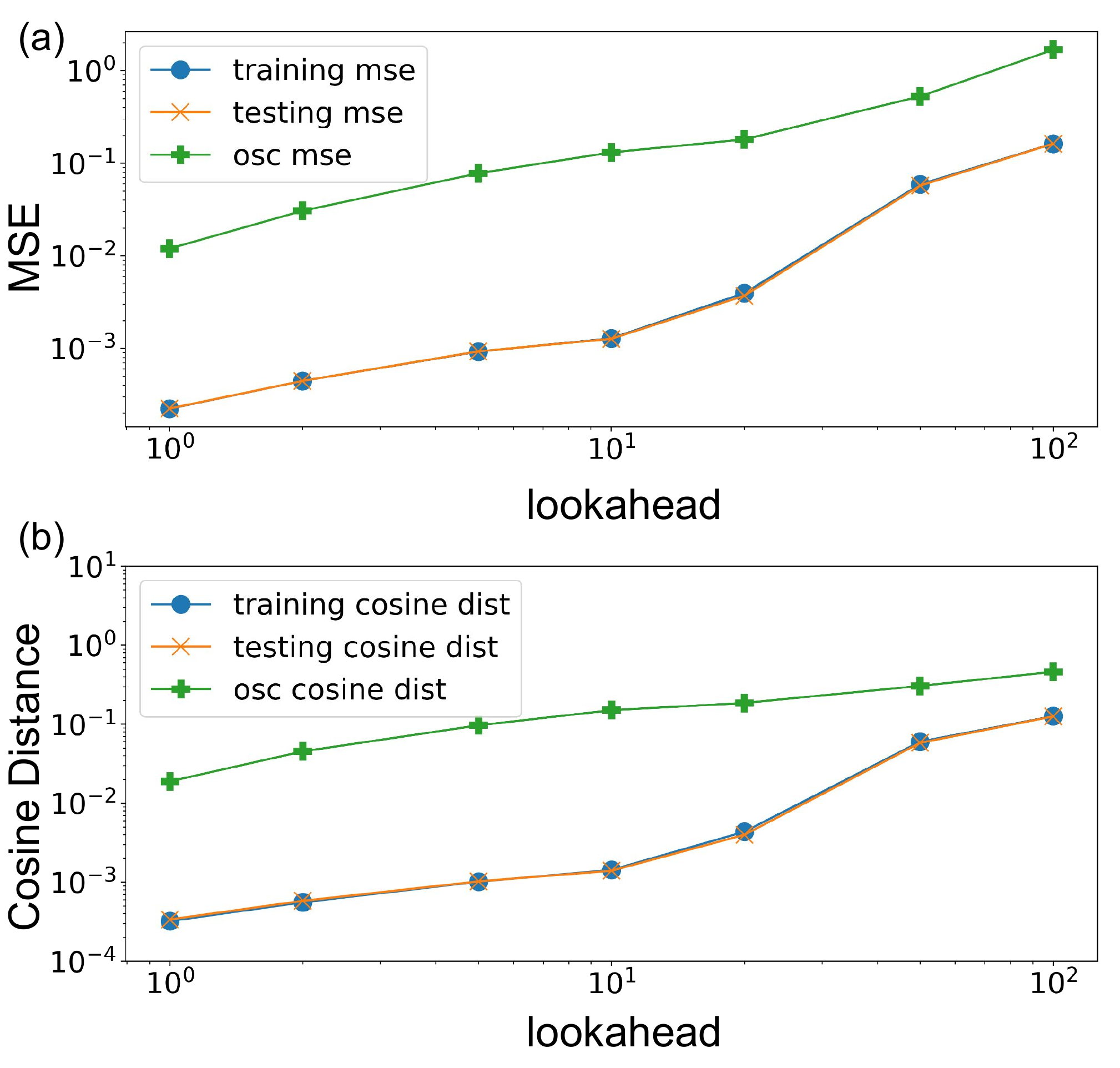}
    \caption{\textbf{Flow field model evaluation.} The plot of (a) mean squared error, and (b) cosine distance, on the training data, testing data, and vortex shedding data with an oscillating cylinder. The lookaheads are 1, 2, 5, 10, 20, 50, and 100.}
    \label{fig:vortex_eval}
\end{figure}
\begin{figure}[h]
    \centering
    \vspace{1 em}
    \includegraphics[width=0.48\textwidth]{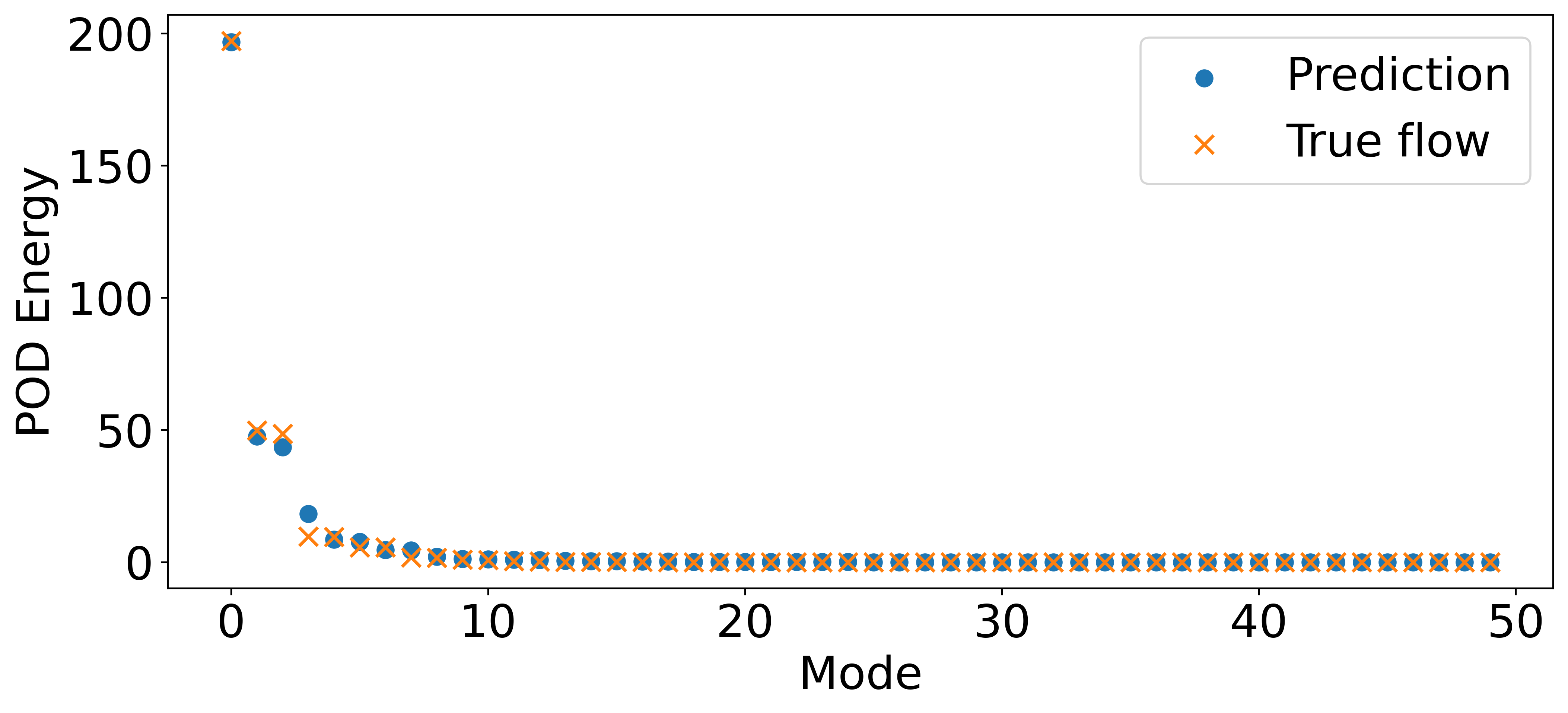}
    \caption{\textbf{POD energy plot:} (a) ground truth flow field, and (b) predicted flow field, from the 300th to the 350th snapshots. It can be observed that the predicted flow field has a similar energy distribution as the ground truth.}
	\vspace{-1em}
    \label{fig:vortex_pod}
\end{figure}

The MSE and cosine similarities are shown in Fig.~\ref{fig:vortex_eval}. Three key observations are to be made. First, it can be observed that the error increases as the the lookahead increases, i.e., the numerical integration is performed for longer time horizons from the initial condition. This is true for solutions of differential equations which can drift overtime as errors accrue. With sampling, our method can correct the drifts by updating the initial conditions used by the model. Second, the model performance on training and testing data are mostly the same for vortex shedding. This is due to the periodic nature of the flow. Having similar errors on the testing data means the learned model has correctly captured the limit cycle exhibited by the flow. Last but not the least, the error on the flow with oscillating cylinder is much higher than if the cylinder is stationary. This indicates a distribution shift when the cylinder is oscillating. The exact values of MSE and cosine similarities are included in the appendix.

Additionally, the learned model can generate a flow field that has a similar energy distribution as the ground truth, as show in Fig.~\ref{fig:vortex_pod}. With the first four most energetic basis, $89.8\%$ and $91.8\%$ of the energy is captured by the predicted and ground truth flow respectively.

Note that although the vortex shedding is periodic, any model based on differential equations cannot avoid having drift when performing long time horizon predictions, as shown in Fig.~\ref{fig:vortex_eval}. Hence it is a good illustrative example of how sampling could help mitigate drift in the solutions. Furthermore, KNODE has been shown to model more complex flows such as the ones that are spatiotemporally chaotic~\cite{Jiahao2021Knowledgebased}, which will be part of the future work.

\begin{figure}[h]
	\centering
    \vspace{1 em}
	\subfigure[]{ \includegraphics[width=0.8\columnwidth]{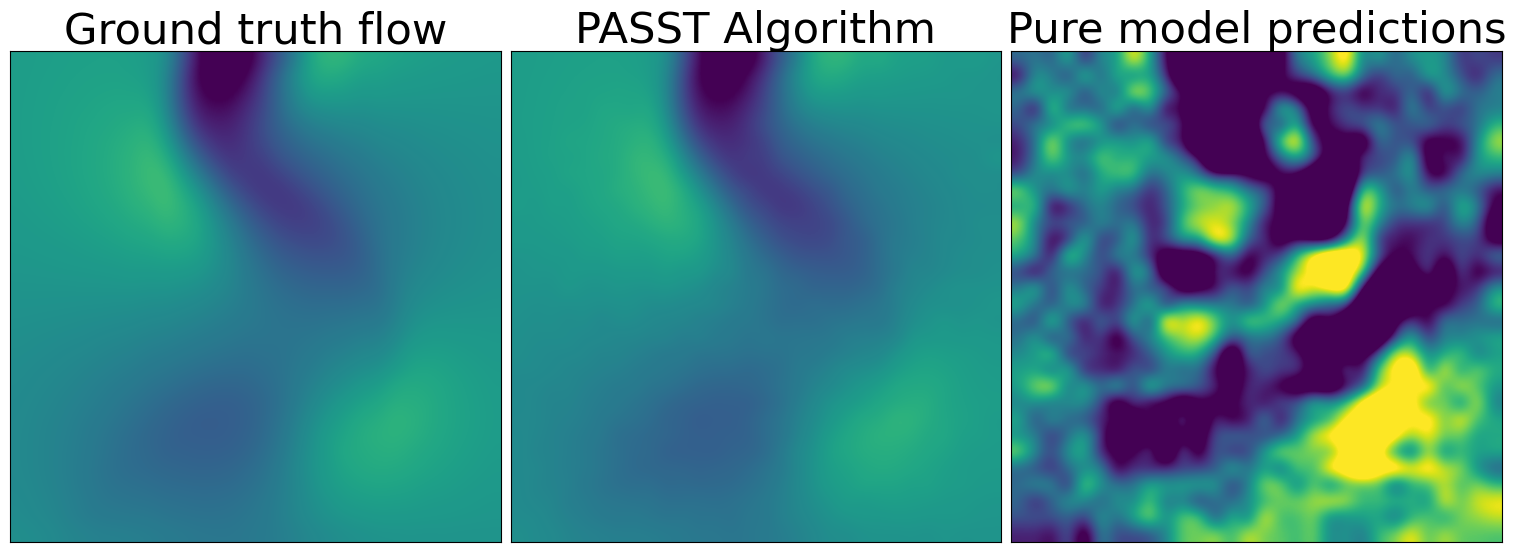} \label{fig:fixed_rmse}}
	\subfigure[]{ \includegraphics[width=0.92\columnwidth]{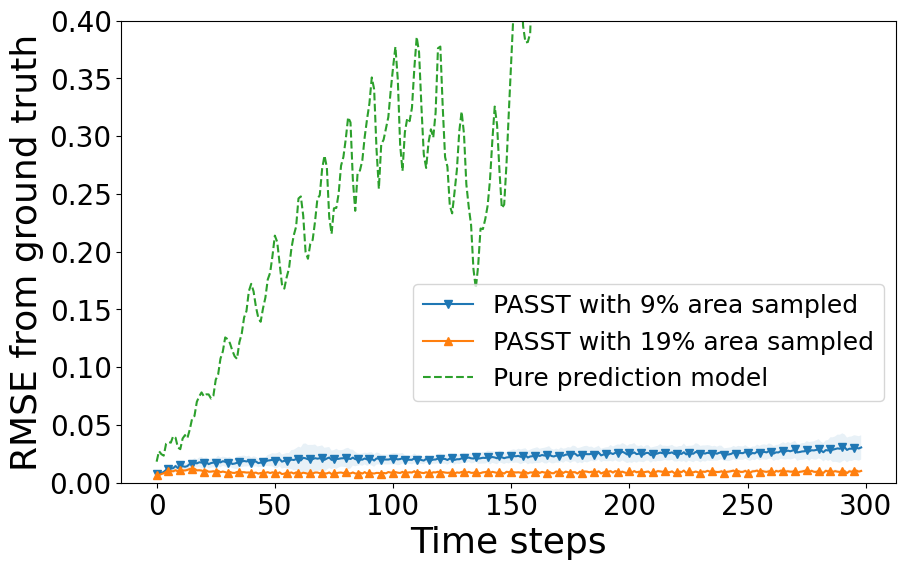} \label{fig:fixed_rmse}}\hspace{-1em}
	\subfigure[]{ \includegraphics[width=0.92\columnwidth]{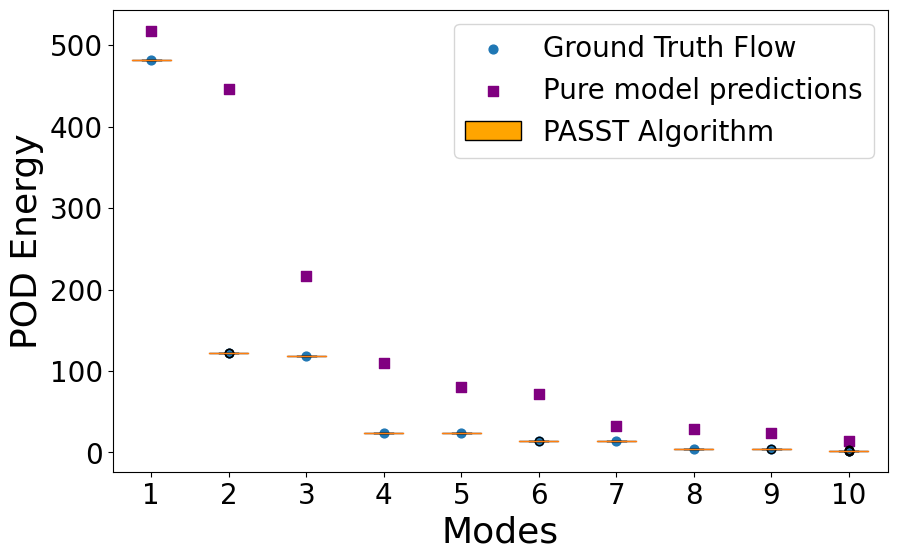} \label{fig:fixed_pod}}\hspace{-1em}
	\vspace{-5pt}
	\caption{(a) Comparison of 2D visualizations of the flow field using PASST and pure model predictions $250$ time steps into the sampling process. (b) Presents RMSE of the reconstructed state with the underlying ground truth field. (c) Presents POD energy plots of the predicted flow field and that of the ground truth flow. It can be observed that the predicted flow field from PASST has a similar energy distribution as the ground truth. The presented results are from 30 trials.}
	\vspace{-1em}
	\label{fig:fixed_cylinder_result}
\end{figure}

\subsection{Evaluation of the \textbf{PASST} Algorithm}\label{subsec:sampling_eval}

The results in the Subsection~\ref{subsec:model_eval} show that the trained predictive model can predict the flow field very well for about a fifty time steps in the future. But, the prediction quality goes on degrading as we start predicting hundreds of steps ahead in time from the initial state. To achieve understanding of a spatiotemporal process persistently, we need to be able to accurately and consistently estimate the evolution of a fluid process over long periods of time. We achieve this by bringing together predictive modeling and adaptive spatial sampling with a robot. We use the short term predictions from the model and plan trajectories that maximize the sampling rewards. Once a set of measurements are collected over a fixed horizon, a field map is generated by fitting the sampled data using a Gaussian Processes, which can be used as the next initial state for the predictive model. Thus the cycle between KNODE predictive model and the policy gradient based sampling keeps achieving a very good revaluation scores even on long-period persistent monitoring.

To begin with, we present the results of evaluating PASST on the first vortex shedding fluid process where the cylinder is static in the center of the field. We compute similar metrics as before --- Fig.~\ref{fig:fixed_rmse} presents the root mean squared error (RMSE) between the ground truth and temporal estimates of the fluid process by PASST algorithm. We tried $6$ different robot path lengths between re-initialization of the initial state to KNODE, thus resulting in different $\%$ of the area covered. For brevity, we present only $2$ representative cases out of the $6$ variations. All the evaluations presented for PASST algorithm are from $30$ trials. The RMSE plot shows that even with a very small appended data from robotic sampling, we are able to achieve persistent monitoring of the fluid process for long. For the qualitative metric, the goal is to evaluate whether the flow generated by the PASST algorithm is ``similar'' to the ground truth. To achieve this, we adopt POD to compare the energy distribution of the estimated flow~\cite{holmes_lumley_berkooz_1996}. In Fig.~\ref{fig:fixed_pod}, it can be observed that the predicted flow field from PASST has a similar energy distribution as the ground truth.

\begin{figure}[h]
	\centering
	\subfigure[]{ \includegraphics[width=0.92\columnwidth]{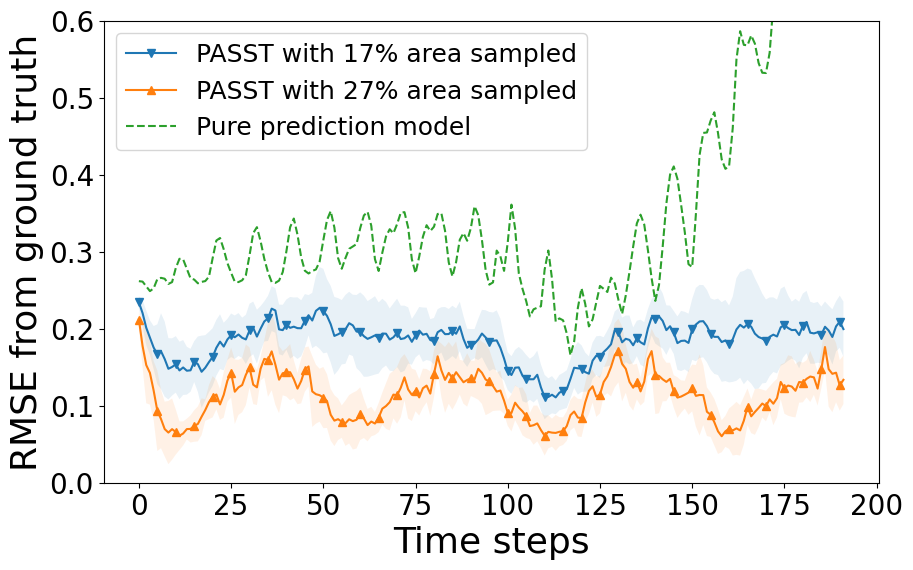} \label{fig:osci_rmse}}\hspace{-1em}
	\subfigure[]{ \includegraphics[width=0.92\columnwidth]{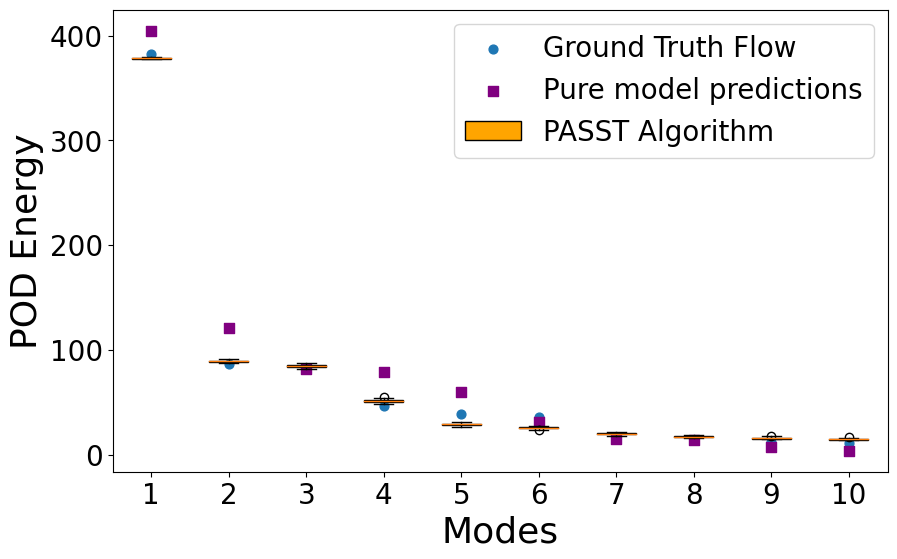} \label{fig:osci_pod}}\hspace{-1em}
	\vspace{-5pt}
	\caption{(a) Presents root mean squared error of the reconstructed state with the underlying ground truth field. (b) Presents POD energy plots of the predicted flow field and that of the ground truth flow. It can be observed that the predicted flow field from PASST has a similar energy distribution as the ground truth. The presented results are from 30 trials.}
	\vspace{-1em}
	\label{fig:oscillating_cylinder_result}
\end{figure}

Fig.~\ref{fig:oscillating_cylinder_result} presents the results of evaluating PASST on the second vortex shedding fluid process where the cylinder is oscillating from top to bottom at a rate of one oscillation per $96$ time steps. In this set of experiments, the KNODE predictive model was only trained on the static cylinder vortex shedding data and the \textbf{PASST} algorithm used predictions coming from this model to sample from oscillating cylinder vortex shedding data. These results show that the proposed \textbf{PASST} algorithm is able to sample efficiently even in the absence of a perfectly trained predictive model. These plots in Fig.~\ref{fig:oscillating_cylinder_result} validate the ability of \textbf{PASST} algorithm to generalize to sample from fluid processes that are not in the training repertoire. Even though the RMSE in Fig.~\ref{fig:osci_rmse} is slightly higher, our algorithm (PASST) is able to persistently monitor the fluid process with a bounded error as presented in these plots. 

\subsection{Experiments with Nowcast Ocean data}\label{subsec:real_data}

\begin{figure}[h]
    \centering
    \vspace{1 em}
    \includegraphics[width=0.45\textwidth]{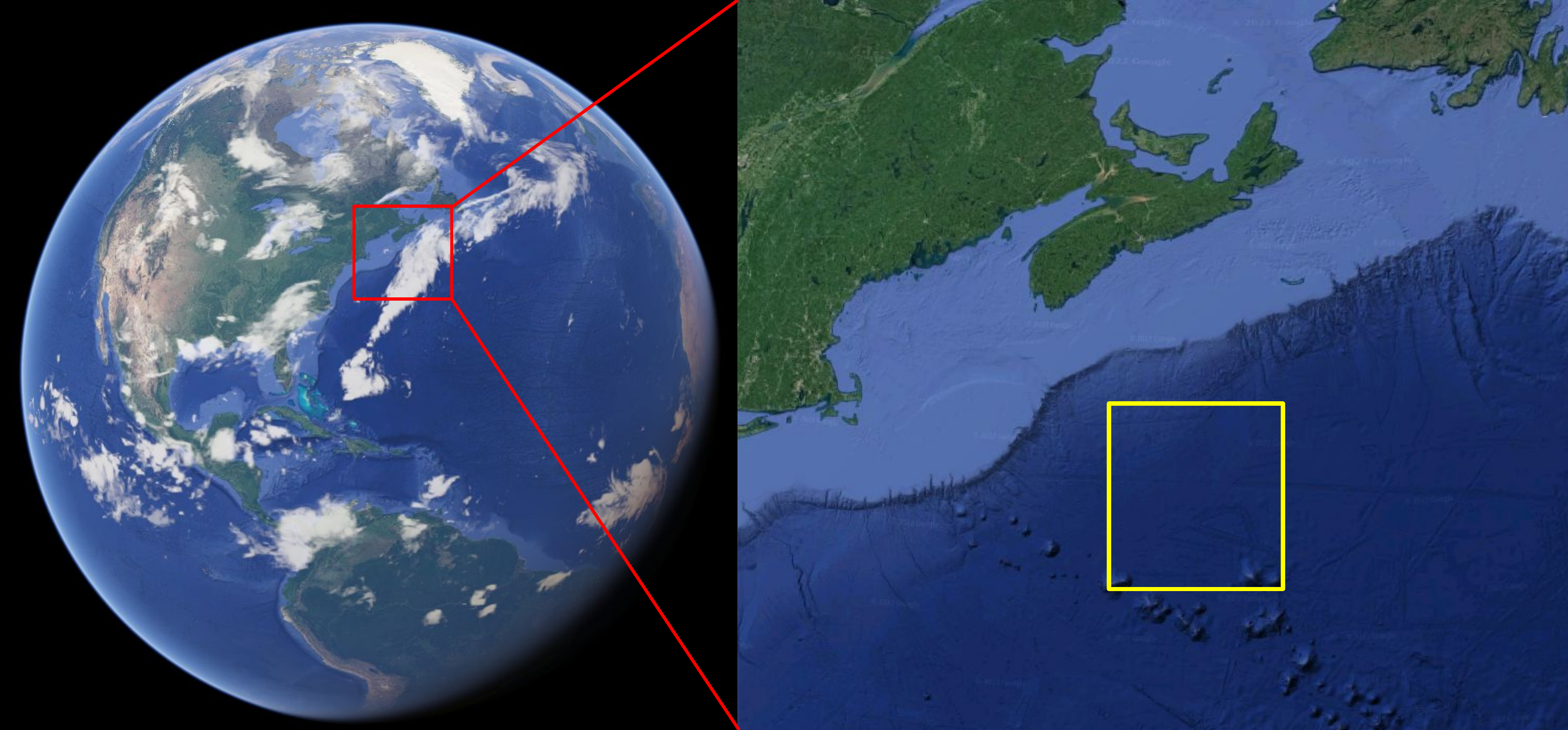}
    \caption{The bounding box for the nowcast data in North Atlantic Ocean.}
    \label{fig:noaa_data_location}
\end{figure}

\begin{figure}[h]
    \centering
    \includegraphics[width=0.48\textwidth]{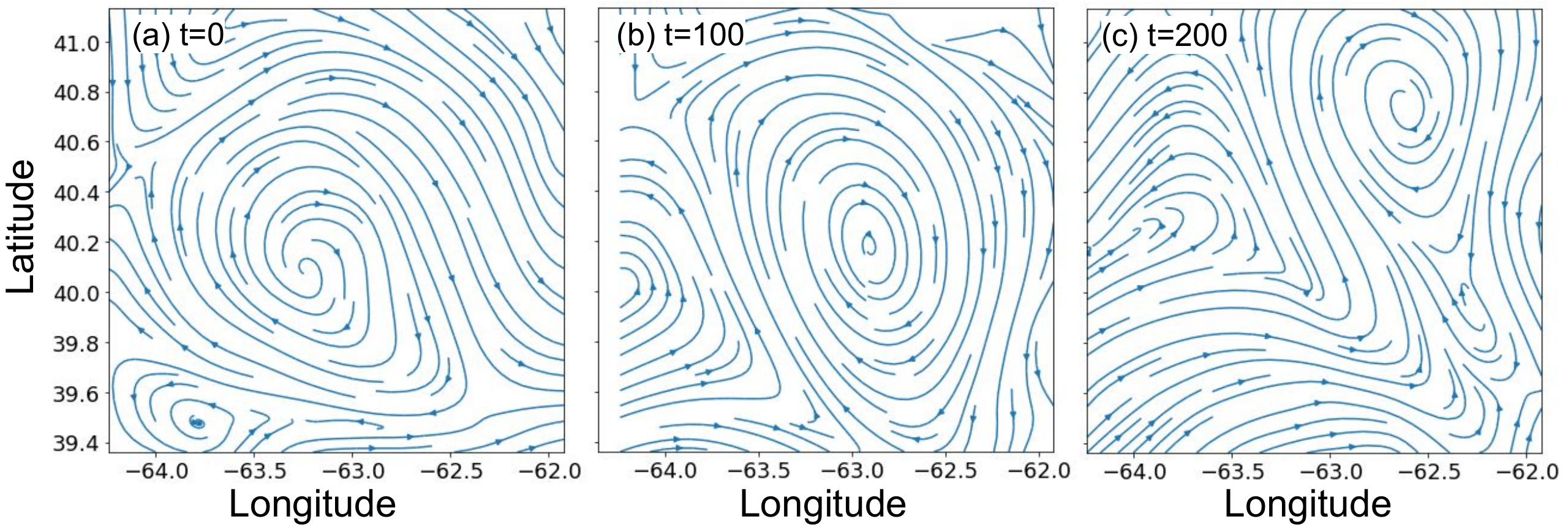}
    \caption{The streamline plot of NOAA data at snapshots (a) t = 0, (b) t = 100, (c) t = 200.}
	\vspace{-1em}
    \label{fig:noaa_data}
\end{figure}

Next, we consider the nowcast data of ocean flow in the North Atlantic Ocean generated by NOAA. The data, hereinafter referred to as NOAA data, is also on a 30 by 30 grid, with an approximate spatial resolution of 6.6km in both longitude and latitude. The area of ocean considered is within the bounding box between the latitudes 39.360065 and 41.130714. and longitudes -64.240000 and -61.920040, as shown in Fig.~\ref{fig:noaa_data_location}. The data is the hourly nowcast for a duration of $21$ days from Nov $2^{nd}$ to Nov $22^{nd}\ 2021$ ($504$ data points). The flow at snapshots $0, 100$, and $200$ are visualized with streamlines in Fig.~\ref{fig:noaa_data}. The model used for training is the same as the one used for vortex shedding. The same architecture as shown in Fig.~\ref{fig: node arch} is adopted.

\begin{figure}[ht]
	\centering
    \vspace{1 em}
	\subfigure[]{ \includegraphics[width=0.8\columnwidth]{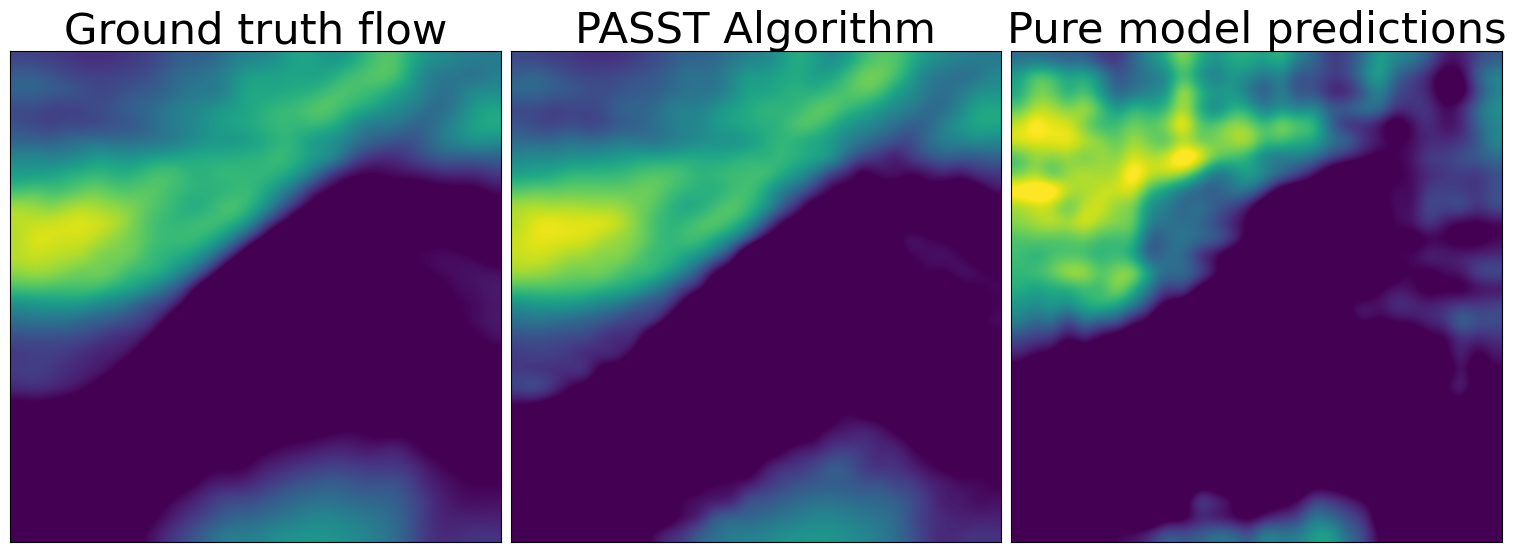} \label{fig:noaa_rmse}}
	\subfigure[]{ \includegraphics[width=0.92\columnwidth]{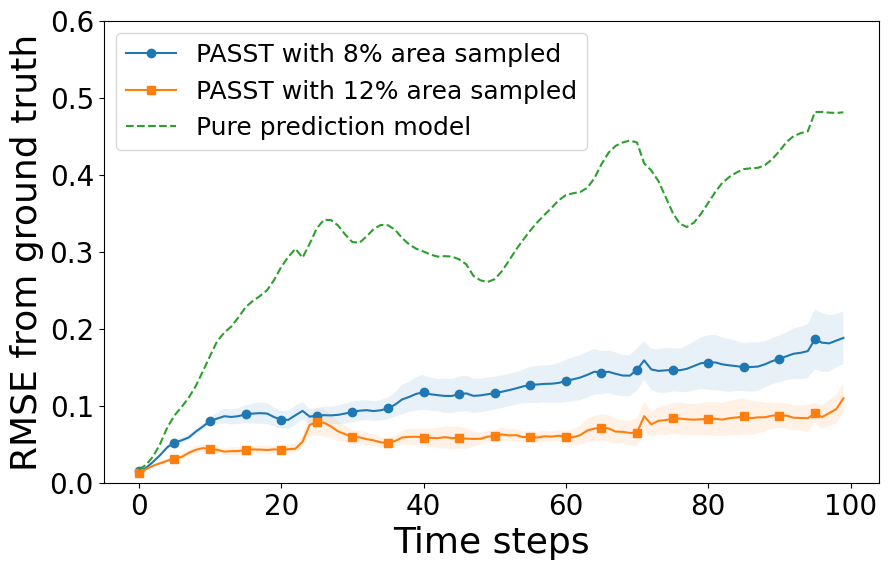} \label{fig:noaa_rmse}}\hspace{-1em}
	\subfigure[]{ \includegraphics[width=0.92\columnwidth]{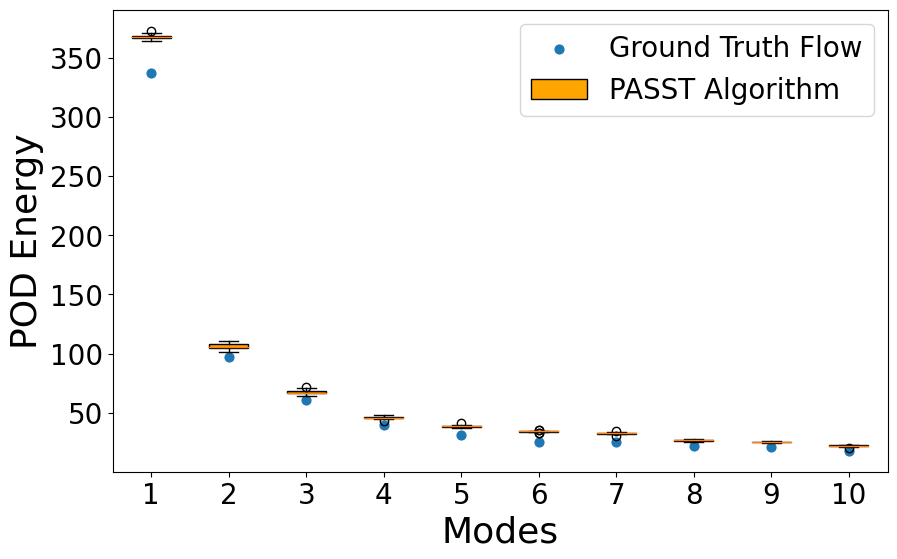} \label{fig:noaa_pod}}\hspace{-1em}
	\vspace{-5pt}
	\caption{\textbf{Nowcast Ocean Data:}(a) Comparison of 2D visualizations of the flow field using PASST and pure model predictions. (b) Presents RMSE of the reconstructed state with the underlying ground truth field. (c) Presents POD energy plots of the predicted flow field and that of the ground truth flow. It can be observed that the predicted flow field from PASST has a similar energy distribution as the ground truth. The presented results are from 30 trials.}
	\vspace{-1em}
	\label{fig:noaa_result}
\end{figure}

Fig.~\ref{fig:noaa_rmse} and Fig.~\ref{fig:noaa_pod} present similar performance metrics as before, RMSE and POD respectively. The RMSE curves for our PASST algorithm do exhibit a very good improvement compared to only predictions from pure model. However, the curves seem to be rising slowly and do not show the property of being bounded as it was in for the other two test cases. One reason could be the amount of area covered between re-initialization of the sampler is small, thus resulting in longer time to plateau. Also, the real-world size of the region considered for these experiments is large, hence increasing the robot path length to achieve further coverage between re-initialization is not very realistic. We do plan to test this hypothesis out in simulation in the near future.

\section{CONCLUSION AND FUTURE WORK}

In this paper we presented an online adaptive sampling algorithm that leverages predictive models to persistently monitor a spatiotemporal fluid process (\textbf{PASST} Algorithm). Our evaluations, on both numerically simulated data and nowcast ocean data, not only highlight the strengths of our approach but also prove the applicability of this approach to real-world data acquisition problems. The performance metrics, RMSE and POD, validate both the quantitative and qualitative aspect of the PASST algorithm to persistently monitor a spamporal process and also reconstruct the fluid flow dynamics accurately. 

We see various future directions for this work. We plan to extend this work to sampling with robot teams, which will enable monitoring of processes at much larger scales than a single robot. This will further bring interesting research questions such as consensus for neural networks and multiagent reinforcement learning in the context of adaptive sampling. Additionally, the predictive model in our method can be replaced by other modeling methods. We will adopt alternative models such as LSTMs and transformers to compare performances. Another potential extension is online learning of predictive models, where we not only update the input to the model but also retrain the model itself during robot deployment. Last but not the least, we plan to carry out real-world experiments in natural bodies of water to validate our algorithm.

\section*{APPENDIX}
\subsection{Training setup}
We use the Euler's method as the numerical solver, and in between each sampled interval, we let the Euler's method take 6 steps. For training the KNODE model, we use the full 100 steps of sampled flow data as one batch for training. For the NOAA data, a total of 400 steps are sampled and it exceeds the GPU memory, and therefore we divide up the data into 3 batches by including the $i^{th}$ snapshot in the $(i \mod 3)^{th}$ batch.

The detailed neural network architecture is shown in Table \ref{tbl: flow_model}.
\begin{table}[H]
\centering
\caption{Neural network architecture for learning the flow fields.}
\label{tbl: flow_model}
\resizebox{0.48\textwidth}{!}{
\begin{tabular}{ccccccc} \toprule\midrule
    {Layer } & { Type } & { Channel Out } & { Kernel Size} & {Stride} & {Padding} & { Activation } \\ \midrule
    1  & Conv & 32 & 3 & 1 & 0 & Tanh\\
    2  & Conv & 64 & 3 & 1 & 0 & Tanh\\
    3  & Conv & 128 & 3 & 2 & 1 & Tanh\\
    4  & Conv & 128 & 3 & 2 & 1 & Tanh\\
    5  & Conv & 128 & 3 & 2 & 0 & Tanh\\
    6  & Linear & 128 & 3 & 2 & 1 & ReLU\\
    7  & Linear & 128 & 3 & 2 & 1 & ReLU
 \\\midrule\bottomrule
\end{tabular}}
\end{table}

\section*{Acknowledgement}
This work was supported by NSF DUE-1839686 and NSF IIS 1910308.

\bibliographystyle{IEEEtran} 
\bibliography{IEEEabrv, refs}

\addtolength{\textheight}{-12cm}
\end{document}